\documentclass{journal}
\usepackage{cite}
\usepackage{amsmath,amssymb,amsfonts}
\usepackage{algorithmic}
\usepackage{graphicx}
\usepackage{textcomp}

\usepackage[utf8]{inputenc} 
\usepackage[T1]{fontenc}    
\usepackage{hyperref}       
\usepackage{url}            
\usepackage{booktabs}   
\usepackage{subcaption}
\usepackage{adjustbox} 
\usepackage{array}
\usepackage{color,soul}

\usepackage{bm}
\usepackage{authblk}

\def\BibTeX{{\rm B\kern-.05em{\sc i\kern-.025em b}\kern-.08em
    T\kern-.1667em\lower.7ex\hbox{E}\kern-.125emX}}

\title{Just In Time Transformers}

\author[1]{Ahmed Ala Eddine Benali}
\author[1]{Massimo Cafaro}
\author[1]{Italo Epicoco}
\author[1]{Marco Pulimeno}
\author[2]{Enrico Junior Schioppa}
\affil[1]{Department of Engineering for Innovation, University of Salento, Lecce, 73100 Italy}
\affil[2]{Inmatica S.p.A. – Research and Development}

\begin{document}

\maketitle

\begin{abstract}
Precise energy load forecasting in residential households is crucial for mitigating carbon emissions and enhancing energy efficiency; indeed, accurate forecasting enables utility companies and policymakers, who advocate sustainable energy practices, to optimize resource utilization. Moreover, smart meters provide valuable information by allowing for granular insights into consumption patterns. Building upon available smart meter data, our study aims to cluster consumers into distinct groups according to their energy usage behaviours, effectively capturing a diverse spectrum of consumption patterns. Next, we design JITtrans (Just In Time transformer), a novel transformer deep learning model that significantly improves energy consumption forecasting accuracy, with respect to traditional forecasting methods. Extensive experimental results validate our claims using proprietary smart meter data. Our findings highlight the potential of advanced predictive technologies to revolutionize energy management and advance sustainable power systems: the development of efficient and eco-friendly energy solutions critically depends on such technologies.
\end{abstract}


\maketitle

\section{Introduction}
\label{sec:introduction}

Electrical energy is one of the main drivers of modern society, powering everything from our homes and industries to the various nuances of our daily routines.With rapid economic growth, rising populations, and a shift toward renewable energy sources, accurately predicting energy consumption has become essential for minimizing carbon emissions and addressing climate change. Representing around 27\% of global electricity demand, the residential sector plays a critical role in this pursuit \cite{guo2022predicting}. In response to these dynamics, the International Energy Agency (IEA) released its "Electricity 2024" report\footnote{https://www.iea.org/reports/electricity-2024/executive-summary}, providing insights into recent trends and future forecasts for global electricity demand. According to the report, global electricity demand grew by 2.2\% in 2023, slightly lower than the 2.4\% observed in 2022, primarily due to reduced consumption in advanced economies. However, demand is expected to accelerate to an average of 3.4\% per year from 2024 to 2026. This expected increase highlights the need for efficient energy management strategies that rely significantly on advancements in forecasting techniques.

Advancements in time-series forecasting has been and continues to be crucial in addressing the complexities of the energy sector. The journey began with the adoption of statistical models (e.g., ARIMA, the Autoregressive Integrated Moving Average) in the 1970s, valued for their simplicity and effectiveness in handling linear time-series data \cite{box2015time}. However, as the complexity and volume of data increased, more sophisticated models became necessary to improve forecasting accuracy. Support Vector Machines (SVM) and Random Forests (RF) were among the first machine learning algorithms used in the late 20th century, enabling the identification of non-linear patterns beyond the reach of traditional models \cite{cortes1995support,breiman2001random}. The development of Recurrent Neural Networks (RNNs) marked a significant advancement, enabling enhanced processing of sequential data by maintaining memory across data sequences, which was critical for forecasting tasks \cite{rumelhart1986learning}. The introduction of Long Short-Term Memory (LSTM) networks addressed the limitations of earlier RNNs by capturing long-term dependencies within datasets \cite{hochreiter1997long}. Convolutional Neural Networks (CNNs), primarily used for image processing, have also been successfully adapted for time-series forecasting, recognizing local patterns within the data \cite{lecun1998gradient}.

More recently, the adaptation of Transformer models, originally designed for Natural Language Processing (NLP) tasks, has revolutionized time-series forecasting. These models employ self-attention mechanisms that adeptly capture both short- and long-term dependencies in data, significantly enhancing forecasting precision compared to earlier models \cite{vaswani2017attention}. Given these advancements, effective energy management can lead to substantial cost savings for consumers, enhance the reliability of power systems, and contribute to a more sustainable future\cite{jaramillo2024adaptive,mathumitha2024intelligent,barbaresi2022application,al2023hybrid}.

The landscape of residential energy consumption is rapidly evolving with the integration of technologies such as electric vehicles, solar panels, and wind turbines. These additions introduce considerable variability and complexity into consumption patterns, challenging traditional forecasting methods \cite{mishra2024decode,al2023hybrid}. For instance, electric vehicles require substantial power for charging, often during off-peak hours \cite{barbaresi2022application}, while households with solar panels may unpredictably feed energy back to the grid, depending on weather conditions. Similarly, wind turbines add another layer of variability, influenced by factors like wind patterns and geographic location \cite{jaramillo2024adaptive}. These technologies not only increase fluctuations in energy usage but also create bidirectional energy flows as households become both consumers and potential energy suppliers \cite{mishra2024decode}. Such shifts demand forecasting models capable of handling real-time, high-frequency data and adapting to rapid changes. Furthermore, smart meters provide continuous streams of granular data, capturing real-time consumption fluctuations. While valuable, this data intensifies the need for adaptive models capable of processing large-scale, complex datasets \cite{mathumitha2024intelligent,liu2023review}. Behavioral and demographic factors, including household size, energy awareness, and appliance usage, further influence consumption patterns, creating diverse profiles that require models to go beyond traditional methods to capture the unique nuances of individual households effectively.

To address the aforementioned challenges in residential energy consumption prediction, we propose the JITtrans (Just In Time Transformer) model, a novel approach leveraging smart meter data to cluster residential energy consumers based on their usage patterns. By forming distinct clusters, we capture diverse consumption trends reflective of various household behaviors and lifestyles. Analyzing a dataset spanning 28 months from an Italian power company, we deploy JITtrans, a transformer model designed to predict energy consumption within these clusters. Moreover, by leveraging a modular design where each transformer in the ensemble is specialized for a different forecast horizon, JITtrans achieves robust generalization for each cluster, minimizing cumulative errors over time. This modular and adaptive approach ensures that JITtrans not only enhances predictive accuracy but also addresses the diverse and evolving energy usage dynamics found in residential settings. Furthermore, extensive experimental results reveal that our JITtrans model outperforms established deep learning models such as LSTM, CNN-LSTM, GRU, and even the basic transformer model. Our contributions are three-fold:

\begin{itemize}
\item We provide a comprehensive review of related work, categorized into key areas that reflect the evolution and current trends in electricity consumption forecasting;
\item We introduce a highly effective approach to improve the predictive accuracy of energy load forecasting, based on clustering the users and a novel transformer model we designed, JITtrans;
\item We evaluate our model generalization ability across clusters featuring varied behavioural patterns in residential energy consumption and provide an extensive comparative analysis of our proposed approach against traditional deep learning models.
\end{itemize}

The remainder of this paper is structured as follows. Section \ref{related} recalls related work in clustering and energy load prediction. Section \ref{problem} details the problem statement, highlighting specific challenges our research aims to address. Section \ref{architecture-sec} presents the overall architecture and the JITtrans model. We discuss the experimental results in Section \ref{results}. Finally, we draw our conclusions in Section \ref{conclusions}, summarizing our findings and suggesting possible future research directions.

\section{Related Work}
\label{related}

A wealth of energy consumption data has become available with the advent of smart meters, facilitating significant advancements in prediction models. This section recalls  related works in the field, categorizing them into key areas that reflect the evolution and current trends in electricity consumption forecasting. The reviewed works span from early developments in utilizing smart meter data to advanced machine learning models, behavioural analytics, and innovative hybrid approaches, showcasing the collective contributions and emerging trends in this domain. 

%

\subsection{Early Developments in Smart Meter Data Utilization}

Gajowniczek and Zabkowski\cite{gajowniczek2014short} were among the first to leverage time series data from smart meters to enhance forecasting models. By employing Artificial Neural Networks (ANNs) and Support Vector Machines (SVMs), they demonstrated the potential of using individual smart meter data for short-term electricity use predictions. This work laid the ground for more sophisticated analyses of energy consumption patterns. Similarly, Ilić et a al. \cite{ilic2013impact} investigated the impact of smart meter data aggregation on forecasting accuracy. Their study emphasized the potential of fine-grained smart meter data to enhance forecast precision by identifying highly predictable consumer groups, which is desirable for efficient energy management and Virtual Power Plants (VPPs). It also highlighted the challenges of predicting individual consumer behaviour and the limitations of random aggregation methodologies.

\subsection{Behavioral and Societal Impact Analysis}

Starting in 2016, Zhou and Yang \cite{zhou2016understanding} investigated the use of big data analytics from smart meters to improve electricity consumption forecasting. Their work integrated insights from energy social science, social informatics, and energy informatics, focusing on household energy use behaviours. They also addressed interdisciplinary methodologies and data privacy challenges, laying a thorough understanding of how household behaviours influence energy usage patterns. In 2019 and 2020 many studies, such as \cite{nakabi2019ann} and\cite{wang2020research}, used predictive modelling methods that focused on using an ANN based model to learn how different customers would react to changes in electricity prices. These studies further advanced the understanding of consumer behaviour in energy consumption, emphasizing the responsiveness to economic incentives. By 2021, Brugger et al. \cite{brugger2021energy} expanded the scope by investigating the impact of emerging societal trends like digitalization, the sharing economy, and increased consumer awareness on future energy demands in European countries. Through their research, they came up with four different energy demand scenarios for 2050. These showed how important policymaking is in shaping these trends and what those changes mean for predicting energy use. This gave them a full picture of how societal factors and technological advances affect predicting energy use.

\subsection{Clustering Techniques for Enhanced Forecasting}

This critical section of research focuses on refining forecasts by identifying typical daily consumption profiles through clustering techniques and integrating behavioral insights. Shahzadeh et al. \cite{shahzadeh2015improving} improved load forecast accuracy by clustering consumers into distinct groups based on detailed smart meter data, training individual neural networks for each cluster, and highlighting the critical role of feature selection in clustering for enhanced forecast acuracy. Similarly, Yildiz et al. \cite{yildiz2018household} leveraged historical smart meter data to refine household electricity load forecasts through a dual method approach that combined the Smart Meter Based Model (SMBM) for short-term prediction with the Cluster-Classify-Forecast (CCF) method, which identifies typical daily consumption profiles. Wen, Zhou, and Yang \cite{wen2019shape} further developed this approach by introducing a clustering method relying on an improved $k$-means algorithm integrated with Principal Component Analysis (PCA) for dimensionality reduction. They employed Dynamic Time Warping (DTW) for effectively grouping similar electricity consumption patterns, thus providing valuable insights for Demand Side Management (DSM) and personalized energy-saving strategies. Extending these methodologies, Czétány et al. \cite{czetany2021development}, and Michalakopoulos et al. \cite{michalakopoulos2024machine} employed a range of clustering techniques, including $k$-means, fuzzy $k$-means, Hierarchical Agglomerative Clustering (HAC), and Density-based Spatial Clustering of Applications with noise (DBSCAN), to develop detailed consumption profiles that categorize nearly a thousand households. Their work emphasizes the impact of factors like meter type, day, seasonality, and housing type on consumption patterns, which is crucial for tailoring Demand Response (DR) programs and improving forecasting accuracy. Both sets of studies demonstrate how important it is to (i) include human behaviour in forecasting models, (ii) identify the best consumer groups, and (iii) use clustering techniques in a broader way to understand how people use energy. Studies like \cite{yang2018model,benali2023smart} have also used profile clustering analysis to look into how customers behave when they use energy. This study shows that behavioural insights are important to improve load forecasting and energy management strategies. These  efforts highlight the critical role of clustering and profiling in understanding consumer behaviour, forecasting electricity consumption accurately, and optimizing energy utilization strategies.

\subsection{Advancements in Deep Learning Models}

The advancements in deep learning models for electricity consumption forecasting emphasize the integration of novel techniques and algorithms to enhance prediction accuracy. Li et al. \cite{li2015building} proposed an optimized ANN model refined with an improved Particle Swarm Optimization (PSO) algorithm and PCA, aiming to improve short-term electricity consumption predictions. In parallel, Chou and Tran \cite{chou2018forecasting} highlighted the superior accuracy of hybrid Machine Learning (ML) models that combine forecasting with optimization techniques, addressing the challenges of data quality in residential energy consumption patterns.

The application of sequence-to-sequence architectures using both Gated Recurrent Unit (GRU) and LSTM models by Sehovac et al. \cite{sehovac2019forecasting} as well as the integration of CNN with LSTM by Lotfipoor et al. \cite{lotfipoor2020short} and Abraham et al. \cite{abraham2022predicting}, marked advancements in handling time series data effectively for short-term energy forecasting. Additionally, Mahjoub et al. \cite{mahjoub2022predicting} who explored LSTM models in comparison to GRU and Drop-GRU configurations, further highlight the versatility of these models in adapting to diverse temporal patterns in energy forecasting. These hybrid models have shown notable improvements over traditional techniques such as ARIMA by achieving better error metrics like Mean Absolute Error (MAE) and Mean Squared Error (MSE).

Furthermore, the development of novel Transformer-based models by Neo Wu et al. \cite{wu2020deep}, including the Temporal Fusion Transformer (TFT) by Bryan Lim et al. \cite{lim2021temporal} and the Informer model by Haoyi Zhou et al. \cite{zhou2021informer}, has introduced self-attention mechanisms that capture complex patterns and long-term dependencies effectively, enhancing both univariate and multivariate forecasting accuracy. These models emphasize the versatility and robustness necessary for managing large datasets from smart meters and improving long sequence predictions through innovative techniques such as ProbSparse self-attention and self-attention distilling.

Additionally, the work by Fekri et al. \cite{fekri2021deep} on an Online Adaptive RNN and  Mahjoub et al. use of LSTM models \cite{mahjoub2022predicting} have demonstrated the adaptability of deep learning approaches to continuously learn and adapt to new data in real-time, addressing challenges such as concept drift and changing consumption patterns. Finally, recent methodologies like the novel discrete deep learning-based methodology for energy consumption classification by Khashei et al. \cite{khashei2023novel}, the Channel Aligned Robust Blend Transformer (CARD) model by Xue Wang et al. \cite{wang2024card}, and the Pathformer by Chen et al. \cite{chen2024pathformer} have each introduced discrete adjustments and multi-scale architectures that improve time series forecasting. These models utilize channel-aligned attention mechanisms, token blend modules, and adaptive pathways to achieve high accuracy and robustness against overfitting and variable dependencies in energy consumption datasets.

Collectively, these advancements highlight a significant shift towards sophisticated, data-driven deep learning frameworks that not only refine the accuracy of short and long-term forecasts, but also incorporate comprehensive behavioural analyses and adaptability to evolving energy consumption patterns.

\subsection{Bayesian and Probabilistic Models}

Advancements in Bayesian and probabilistic models have significantly improved electricity consumption forecasting using smart meter data. Taieb et al. \cite{taieb2016forecasting} developed an additive quantile regression model for forecasting the distribution of future electricity demand at the household level, addressing the volatility of smart meter data. Concurrently, Wang et al. \cite{wang2018random} enhanced hourly energy predictions in educational buildings using a Random Forest algorithm, which outperformed traditional models like Regression Tree and Support Vector Regression. Furthering this, another study by Wang et al. \cite{wang2018combining} introduced Constrained Quantile Regression Averaging (CQRA) to optimally combine probabilistic forecasts and minimize pinball loss, demonstrating effectiveness with datasets from ISO New England and Irish smart meters. In \cite{weeraddana2021energy}, Weeraddana et al.  used an approach based on a stacked nonparametric Bayesian model with Gaussian Processes to tackle non-linear and non-stationary consumption patterns, surpassing traditional methods like ARIMA and RF in handling short time series data. Additionally, Xu et al. \cite{xu2021clustering} explored monthly residential electricity consumption using a clustering-based probability distribution model, further illustrating the potential of probabilistic models in robust electricity consumption forecasting. These developments highlight a move towards forecasting models that integrate diverse data sources to enhance prediction accuracy and robustness.

\subsection{Hybrid and Ensemble Models}

The development of hybrid and ensemble models represents a novel shift. Researchers have embraced these approaches, combining advanced machine learning techniques to significantly enhance prediction accuracy and effectively address the complexities inherent in data characteristics. Tian et al. \cite{tian2019data} introduced a pioneering method using Generative Adversarial Network (GAN) to forecast energy consumption, which improves prediction by combining original and synthetically generated data, thus addressing data scarcity and enhancing forecast accuracy. Similarly, Alhussein et al. \cite{alhussein2020hybrid} developed a hybrid model that combines CNN for feature extraction and noise filtering with LSTM for capturing temporal dependencies. This model demonstrated superior accuracy over traditional LSTM models by achieving lower Mean Absolute Percentage Error (MAPE), particularly in predicting individual household loads.

Furthering the use of hybrid approaches, Haihong Bian \cite{bian2020study} enhanced short-term electricity consumption forecasting by integrating $k$-means clustering and Fuzzy $C$-Means (FCM) with a neural network. This method clusters users based on consumption patterns before forecasting. Similarly, Khan et al. \cite{khan2021ensemble} developed an ensemble forecasting model that combines spatial-temporal clustering with LSTM and GRU techniques, utilizing high-resolution smart meter data to enhance prediction accuracy in residential settings.

Moreover, Dai and Meng \cite{dai2022online} introduced a model that integrates what they call online model-based functional clustering with functional deep learning, employing a dynamic clustering mechanism and an adaptive functional deep neural network to adapt to time-varying consumption patterns and improve daily load demand forecasts. Adding to the diversity of hybrid models, Saad Saoud et al. \cite{saoud2022household} proposed a forecasting model that merges the Stationary Wavelet Transform (SWT) with transformers. This method decomposes consumption data into high and low-frequency components, allowing to effectively learn and predict complex patterns. Demonstrated to surpass existing models in prediction accuracy and robustness, their approach also exhibits resilience to noise and disturbances.

\subsection{Emerging Trends and Big Data Integration}

In exploring the intersection of emerging trends and big data integration in energy consumption forecasting, researchers are harnessing sophisticated data analytics to enhance accuracy and manage complex datasets effectively. Kalksma et al. \cite{kalksma2018mining} utilized modified Support-Pruned Markov Models (SPMM) to predict appliance usage and energy footprints with high accuracy, showcasing the impact of sequential pattern mining in optimizing energy management despite its data-intensive demands. Similarly, Oprea and Băra \cite{oprea2019machine} leveraged scalable big data frameworks integrated with machine learning algorithms for effective short-term load forecasting in residential settings. Their approach not only improves forecast accuracy but also adapts to changing conditions by processing data from smart meters and weather sensors using a NoSQL database.

These examples illustrate a broader trend in the field: the integration of Artificial Intelligence of Things (AIoT) for real-time data processing \cite{alaba2024aiot}, the application of edge computing to reduce latency, the use of digital twins for scenario simulation, and blockchain for secure energy transactions. Additionally, developments in machine learning models, including deep and reinforcement learning, and integrative data platforms are enhancing the predictive capabilities of energy systems, while predictive maintenance and regulatory analytics ensure optimal and compliant operations. These advancements highlight a significant shift towards sophisticated hybrid and ensemble frameworks that leverage diverse data sources and modelling techniques, refining the accuracy and adaptability of forecasts in smart grid systems.

\subsection{Collective Contributions and Emerging Trends}

All of this related work showcases notable progress in the domain of energy consumption prediction through the utilization of data from smart meters. Initial research demonstrated the viability of utilizing smart meter data for near-term forecasts, but following investigations used clustering methods, other machine learning models, and hybrid approaches to improve precision and dependability. Current developments exploit the availability of high-resolution data by employing innovative techniques, while also highlighting the flexibility of models to adjust to different circumstances. These contributions highlight the continuous development and improvement of forecasting methods, which provide more effective and environmentally friendly energy management practices.

\section{Problem Statement} 
\label{problem}

\subsection{Challenges}

Accurately forecasting electricity consumption at the household level has become increasingly challenging, owing to the dynamic and complex nature of modern energy usage. Based on the findings from the literature review, traditional forecasting methodologies are becoming less effective as they struggle to incorporate the impact of new technologies like electric vehicles, which often charge overnight, and residential solar panels, which introduce significant variability in energy production and consumption. These challenges are further complicated by a myriad of factors including, but not being limited to,  household demographics, appliance usage patterns, meteorological conditions and economic variabilities, each adding layers of complexity to energy management.

Moreover, existing forecasting models whilst proficient at aggregating data for broad community-level analysis, consistently underperform when tasked with applying these insights to individual households. This shortfall primarily stems from two interconnected issues: the models' inability to accurately manage the abrupt daily variations in energy usage and their failure to develop tailored forecasting solutions for each consumer, given the considerable diversity in behaviour and consumption patterns across households. Noise plays a crucial role, since measurements of individual households may have unpredictable behaviour, which averages out when considering, instead, broad-community level data. 

These limitations are exacerbated as the granularity of data collection increases through advanced metering technologies, which unmask the inadequacies of traditional models in capturing the nuanced and varied energy consumption dynamics unique to each household.

\subsection{Methodological Innovations}

\begin{figure*}[h!]
    \centering
    \includegraphics[width=1\textwidth, height=0.4\textheight]{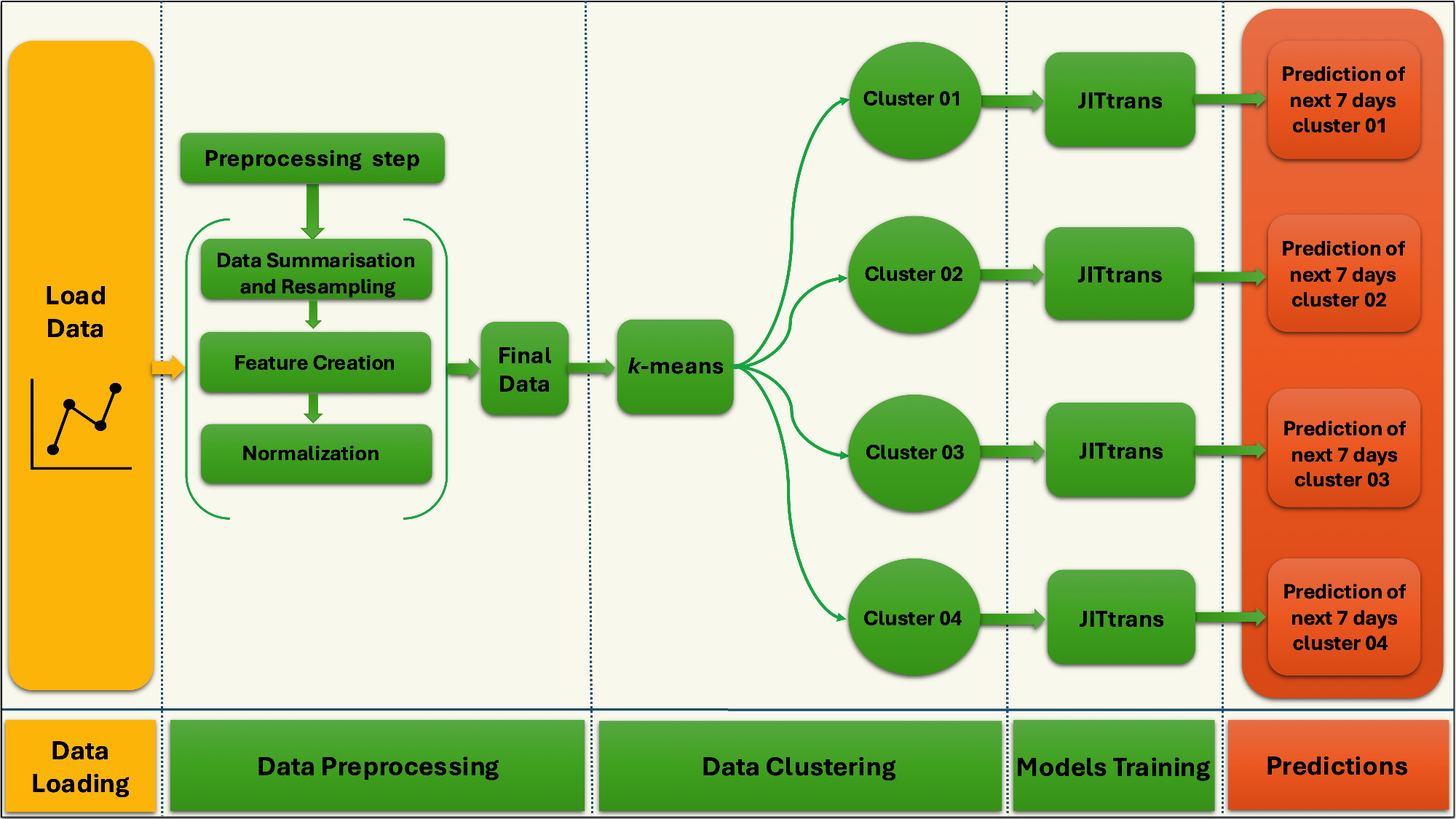} 
    \caption{Overview of the proposed forecasting methodology.}
    \label{overview}
\end{figure*}

To address these deficiencies, this study introduces a novel forecasting methodology that leverages $k$-means clustering to segment the consumer base into distinct groups based on their energy consumption behaviours observed throughout the years. This method does not merely aggregate consumers based on superficial similarities but delves deeper into consistent behavioural patterns, identifying clusters of users who exhibit virtually identical energy usage profiles over extended periods. By pinpointing these distinct clusters, our study leverages the capabilities of transformer models, renowned for their proficiency in handling sequential time-series data. These models are adept at capturing the complex temporal dynamics within each cluster, thus enabling us to apply a specific forecasting approach tailored to the nuanced patterns of each group. This method significantly enhances the accuracy of our forecasts. Our approach is detailed in Figure \ref{overview}, which outlines the progression from data loading to the application of the JITtrans model.

Our innovative approach not only improves the granularity and accuracy of energy consumption forecasts but also supports utility providers in managing demand surges and integrating renewable energy sources more effectively. By refining forecasting accuracy for specific clusters of households, this research contributes significantly to the development of adaptive, efficient, and sustainable energy systems.

This study fills a gap in the existing literature by providing a scalable and practical solution to a pressing global challenge, marking an advancement in the field of energy forecasting. The following section, details the technical specifications and operational framework of the JITtrans model utilized in our forecasting methodology.

\section{Proposed Model Architecture and Forecasting  Strategy} 
\label{architecture-sec}

The main contribution of this work lies in the development of a modular forecasting system using a set of transformer models, each optimized for different forecast horizons. This approach differs from traditional methods that use a single model to forecast over a fixed or variable horizon. Instead, each model within the ensemble is trained with a distinct length of the decoder input, specifically designed to predict the energy consumption at different lead time. This allows the system to specialize in capturing temporal dynamics unique to different forecast ranges, thereby improving accuracy and robustness compared to a one-size-fits-all model. 

During the prediction phase, the JITtrans model leverages a sequential and accumulative prediction strategy. Starting from the shortest horizon model, predictions are generated and used as inputs to the next model in the sequence. This cumulative approach refines the forecast at each step by integrating the predictions from all previous models, thus enhancing the overall forecast as the horizon extends. 

We begin by recalling the technical details related to $k$-means and transformers, then we introduce our modular forecasting system.

\subsection{$k$-means}

Clustering is a technique used to group a set of objects in such a way that objects in the same group (called a cluster) are more similar to each other than to those in other groups. It's a method of unsupervised learning, and it's particularly useful in identifying distinct patterns from large datasets without prior labelling of the data. In the context of electricity consumption, clustering algorithms can identify and group similar usage patterns across different consumers. This is particularly useful for utility companies to tailor customer engagement strategies and optimize the generation and distribution of power.

Representative-based clustering seeks to partition a dataset $D$, consisting of $n$ points $\mathbf{x}_i$, into $k$ clusters (${C = \{C_1, C_2, \ldots, C_k\}}$) within a $d$-dimensional space. Each cluster $C_i$ is characterized by a representative point, a.k.a. the centroid $\mu_i$, which is the mean of all points in the cluster.

\begin{equation}
	\mathbf{\mu}_i = \frac{1}{n_i} \sum_{\mathbf{x} \in C_i} \mathbf{x},
\end{equation}

where \( n_i = |C_i| \) denotes the number of points in cluster \( C_i \).

The objective of clustering is to minimize the within-cluster variance, formulated as:

\begin{align}
    \mathcal{L} = 
     \sum_{i=1}^k \sum_{\mathbf{x} \in C_i} \|\mathbf{x} - \mathbf{\mu}_i\|^2,
\end{align}

 A brute-force approach to determine an optimal clustering involves generating all possible partitions of the $n$ points into $k$ clusters, evaluating their optimization scores, and selecting the partition with the best score. The exact number of ways to partition $n$ points into $k$ non-empty, disjoint subsets is given by the Stirling numbers of the second kind, which gives a total number of unique clustering in the order of $O\left(\frac{k^n}{k!}\right)$. Clearly, counting and evaluating all these clusterings is impractical. Therefore, more efficient algorithms such as $k$-means and expectation-maximization are used in practice for representative-based clustering.

$k$-means \cite{jain1988algorithms} is an iterative algorithm performing two main steps:

\begin{enumerate}
    \item \textbf{Assignment Step:} Each data point is assigned to the nearest cluster centroid based on the Euclidean distance;
    \item \textbf{Update Step:} The centroids of the clusters are updated (they are computed as the mean of all data points assigned to their respective clusters).
\end{enumerate}

These steps are repeated until convergence, which is typically achieved when the centroids no longer change significantly between iterations. The algorithm may converge to a local minimum, and hence it is often run multiple times with different initializations to find the best clustering \cite{hartigan1979algorithm}.

$k$-means is computationally efficient and suitable for large datasets, although the quality of the clustering depends on the initial placement of the centroids. The time complexity of the algorithm is \( O(tnkd) \), where \( n \) is the number of data points, \( k \) is the number of clusters, \( d \) is the number of dimensions, and \( t \) is the number of iterations \cite{zaki2014data}.

\subsection{Transformers}

Transformers are neural network architectures designed to handle sequential data efficiently across various applications. Unlike traditional models that process sequences of data in sequential order, transformers utilize the positional embedding in order to process all \emph{tokens} in the sequence simultaneously; tokens are discrete units representing the items in the sequence, they can be patches of an image, words in a sentence or numerical values in a time series. This parallel processing capability significantly enhances the scalability and efficiency of the model with respect to sequence length and complexity.
The transformer processes the input data through a sequence matrix $\mathbf{X}$, which undergoes transformations via self-attention mechanisms. The aim of self-attention mechanism is to obtain a finer representation of the current token by means of a weighted average of the most relevant (i.e. the most correlated with the current one) tokens in the sequence. The mechanism is based on the use of so-called \emph{queries} $\mathbf{Q}$, \emph{keys} $\mathbf{K}$ and \emph{values} $\mathbf{V}$ matrices, which are different representations of the input tokens, $\mathbf{X}$. Namely, for each token in $\mathbf{Q}$ we compute its correlation with all the tokens in $\mathbf{K}$ as result of the dot product $\mathbf{QK}$; this correlation is then used to obtain the finer representation of the current token as a weighted sum of the tokens in $\mathbf{V}$. $\mathbf{Q}$, $\mathbf{K}$, and $\mathbf{V}$ are derived from the input matrix $\mathbf{X}$ by using different learnable weights:
    
    \begin{equation}
    	 \mathbf{Q} = \mathbf{X}\mathbf{W}^Q, \quad \mathbf{K} = \mathbf{X}\mathbf{W}^K, \quad \mathbf{V} = \mathbf{X}\mathbf{W}^V.
    \end{equation}
  
Where $\mathbf{W}^Q$, $\mathbf{W}^K$ and $\mathbf{W}^V$ are learnable weights specific to queries, keys, and values, respectively. These weights matrices project each input token into a representation according to its role as a key, query, or value.

The self-attention mechanism adjusts the attention each token receives based on the query-key affinity and then uses this attention to form a weighted sum of value vectors. The computation of self-attention requires determining first the attention weights between each pair of input tokens $\mathbf{x}_n, \mathbf{x}_m$:

 \begin{equation}
    	a\left[\mathbf{x}_n, \mathbf{x}_m\right] = \operatorname{softmax}_m\left(\frac{\mathbf{k}_m^T \mathbf{q}_n}{\sqrt{d_k}}\right) := \frac{\exp \left(\frac{\mathbf{k}_m^T \mathbf{q}_n}{\sqrt{d_k}}\right)}{\sum_{m^{\prime}=1}^N \exp \left(\frac{\mathbf{k}_{m^{\prime}}^T \mathbf{q}_n}{\sqrt{d_k}}\right)}.
    \end{equation}

The softmax function ensures that the attention scores, representing the influence weights, are non-negative and sum to 1. The scaling factor $\sqrt{d_k}$, where $d_k$ is the dimension of the token space, is used to help stabilizing the gradients during training.  Self-attention is then computed as 

\begin{equation}
\mathbf{S a}(\mathbf{X})=\mathbf{V} \cdot \operatorname{softmax}\left(\frac{\mathbf{K^T} \mathbf{Q}}{\sqrt{d_k}}\right).
\end{equation}

\subsubsection{Vanilla Transformer}

The vanilla transformer model, introduced by Vaswani et al. \cite{vaswani2017attention}, may be based on an encoder, a decoder or both. Each of these components consists of multiple identical layers. An encoder layer includes a multi-head self-attention mechanism\footnote{A single self-attention mechanism is not enough for a transformer to work correctly, therefore multiple attention mechanisms are used; this is called multi-head self attention.} followed by a position-wise feed-forward network. In contrast, each decoder layer incorporates an additional cross-attention mechanism (i.e., the decoder embeddings attend to the encoder embeddings) between the multi-head self-attention and the feed-forward network. This dual-attention mechanism allows the model to effectively capture and utilize information from the entire input sequence \cite{vaswani2017attention}.

\subsubsection{Input Encoding and Positional Encoding}

To retain the sequential order information, the transformer model employs positional encoding added to the input embeddings. This approach is crucial as the self-attention mechanism itself is permutation invariant, meaning it does not inherently capture the order of the sequence.

\begin{enumerate}

    \item \textbf{Absolute Positional Encoding:} In the vanilla Transformer, positional encoding vectors are generated using sine and cosine functions of different frequencies for each dimension, as defined by the following equations:
   
    \begin{equation}
    	PE_{(pos, i)} = 
    \begin{cases} 
    \sin \left( \frac{pos}{10000^{\frac{i}{d}}} \right) & \text{if } i \% 2 = 0 \\

    \cos \left( \frac{pos}{10000^{\frac{i}{d}}} \right) & \text{if } i \% 2 = 1 
    \end{cases},
    \end{equation} 

    where \( pos \) is the position, \( i \) is the dimension, and \( d \) is the total number of dimensions. This encoding allows the model to incorporate information about the position of each token in the sequence, enabling it to distinguish between tokens in different positions.

    \item \textbf{Relative Positional Encoding:} Some methods focus on pairwise positional relationships between input elements, adding learnable relative positional embeddings to the keys of the attention mechanism \cite{shaw2018self}. There are also hybrid approaches that combine both absolute and relative positional encodings \cite{ke2020rethinking}.
\end{enumerate}

\subsubsection{Multi-head Attention}
The multi-head attention mechanism is a cornerstone of the transformer architecture, enhancing the model's ability to focus on different parts of the input sequence simultaneously. This mechanism employs multiple attention heads, each learning to attend to different parts of the input.
Multi-head attention extends the self-attention mechanism by using multiple sets of learned projections:
\begin{equation}
	\text{MultiHeadAttn}(Q, K, V) := \text{Concat}(T_1, \ldots, T_h)W^O,
\end{equation}
where each attention head $T_i$ is computed as:
\begin{equation}
T_i = \text{Attention}(QW_i^Q, KW_i^K, VW_i^V).
\end{equation}
In this formulation, \( W_i^Q \), \( W_i^K \), and \( W_i^V \) are the learnable weight matrices for the \( i \)-th head, and \( W^O \) is a matrix of learnable weights that projects the concatenated outputs of the multiple attention heads back to the original space. This structure allows the model to jointly attend to information from different representation subspaces at different positions, significantly enhancing its learning capacity.

\subsubsection{Feed-forward and Residual Network}

The feed-forward network in a transformer layer is a fully connected network applied to each position separately and identically:

\begin{equation}
	\text{FFN}(H_0) := \text{ReLU}(H_0W_1 + b_1)W_2 + b_2,
\end{equation}

where \(H_0\) is the output of the previous layer, \(W_1 \in \mathbb{R}^{D_m \times D_f}\), \(W_2 \in \mathbb{R}^{D_f \times D_m}\), \(b_1 \in \mathbb{R}^{D_f}\), and \(b_2 \in \mathbb{R}^{D_m}\) are trainable parameters. Each layer in a transformer also includes residual connections and layer normalization to ensure better gradient flow and faster convergence:

\begin{equation}
	H_0 = \text{LayerNorm}(\text{Attention}(Q,K,V) + X),
\end{equation}

\begin{equation}
	H = \text{LayerNorm}(\text{FFN}(H_0) + H_0),
\end{equation}

where \(\text{LayerNorm}(\cdot)\) denotes layer normalization.

\subsubsection{Recent Innovations in Transformer Models}

The transformer model, introduced by Vaswani et al. \cite{vaswani2017attention}, represents a groundbreaking advancement in sequence modelling, utilizing an encoder-decoder framework. This model has achieved exceptional results across various fields, including NLP \cite{devlin2018bert}, speech recognition \cite{dong2018speech} and computer vision \cite{liu2021swin}. More recently, transformer-based models have become increasingly popular for time series analysis \cite{wen2022transformers}. 

%
In 2024, several new transformer models have emerged, demonstrating promising results in time series forecasting. These models include:

\begin{itemize}
    \item \textbf{CARD} \cite{wang2024card}: Employs channel alignment to robustly enhance forecasting accuracy;
    \item \textbf{Pathformer} \cite{chen2024pathformer}: Utilizes adaptive pathways to efficiently manage multi-scale data;
    \item \textbf{GAFormer} \cite{xiao2024gaformer}: Enhances time series understanding through group-aware embeddings;
    \item \textbf{Transformer-Modulated Diffusion Models for Probabilistic Multivariate Time Series Forecasting} \cite{li2024transformermodulated}: Combines Transformer architectures with diffusion models for probabilistic forecasting;
    \item \textbf{iTransformer} \cite{liu2023itransformer}: Proposes an inverted Transformer architecture for better forecasting performance.

\end{itemize}

These advancements illustrate the continued evolution and adaptation of transformer models to various complex tasks in time series forecasting.

\subsection{Model Design and Functionality}

We now introduce the design of our JITtrans transformer model. The architecture of the proposed ensemble includes multiple transformer models, each specifically trained to predict a distinct day within a 7-day forecast horizon. This setup leverages an input sequence (denoted by $x$) shared among the encoders of each transformer and an input sequence (denoted by $d$) for each decoder. The input sequence to the decoder is progressively extended in size moving from one transformer to the next one, adding the forecasted value obtained by the previous transformer. In our configuration, each transformer model in JITtrans uses a sequence of 30 days as input to each encoder (from \( x_{t-29} \) to \( x_t \)), providing the necessary historical context for accurate forecasting. The decoder input and the targets for each model are structured to sequentially extend the forecast range by concatenating predictions from earlier models.

Denoting by \( \hat{y}_{t+k}^{(i)} \) the forecast for day \( t+k \) of the \( i \)-th model, the first model utilizes a decoder input of 7 days from \( d_t \) to \( d_{t+6} \), and it is  trained to output days from \( d_{t+1} \) to \( d_{t+6} \) and day \( \hat{y}_{t+7}^{(1)} \), which is a new prediction, not included in the decoder input.

Each subsequent model extends the input to its decoder by adding the day forecasted by the previous model. For example, Model 2 uses a sequence of 8 days as input to the decoder (from \( d_t \) to \( d_{t+6} \) and \( \hat{y}_{t+7}^{(1)} \)), incorporating the forecast \( \hat{y}_{t+7}^{(1)} \) predicted by Model 1. Model 2 then makes predictions \( \hat{y}_{t+7}^{(2)} \) and \( \hat{y}_{t+8}^{(2)} \), with \( \hat{y}_{t+8}^{(2)} \) being a new day added to the forecast and \( \hat{y}_{t+7}^{(2)} \) being an updated forecast of day  \( \hat{y}_{t+7}^{(1)} \) based on additional input data.

\begin{figure*}[h!]
    \centering
    \includegraphics[width=\textwidth, height=0.35\textheight]{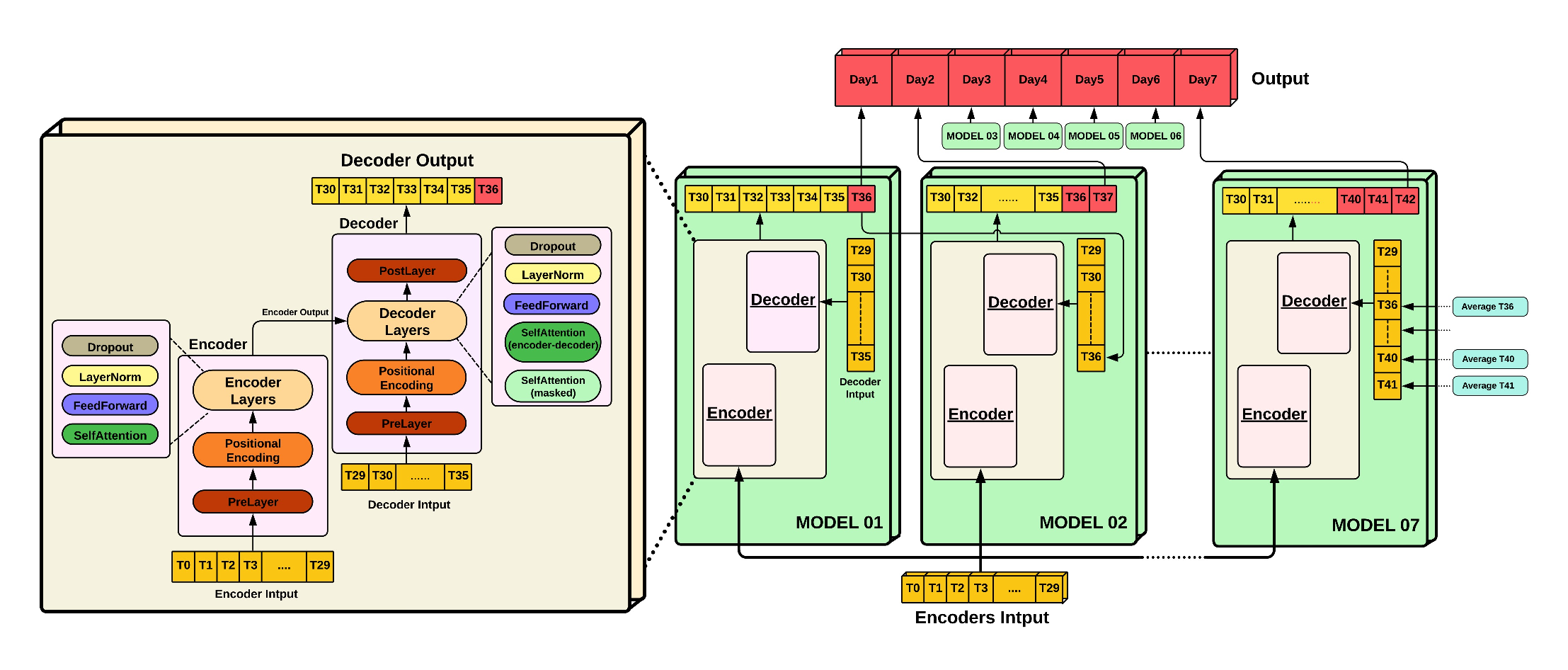} 
    \caption{JITtrans Model architecture.}
    \label{architecture}
\end{figure*}

Figure \ref{architecture} provides a schematic representation of the JITtrans model architecture, illustrating how data flows through the encoder and decoder inputs across the ensemble, thereby clarifying the model's operational framework.

\subsection{Integration and Averaging of Model Outputs}

The ensemble uses the following strategy to integrate and refine predictions: each model contributes to the forecast of subsequent days by utilizing a combination of direct input from its decoder and the averaged predictions from previous models. For instance, Model 3, when predicting \( \hat{y}_{t+9} \), will use the day \( \hat{y}_{t+8} \) directly from Model 2 and take an average of the predictions for \( \hat{y}_{t+7} \) from both Model 1 and Model 2. This averaging mechanism is formally expressed as:

\begin{equation}
	\bar{y}_{t+k}^{(j)} = \frac{1}{j-k} \sum_{i=k}^{j-1} \hat{y}_{t+k}^{(i)},
\end{equation}

where \( \bar{y}_{t+k}^{(j)} \) denotes the averaged value used as input to $j$-th model for day \( t+k \) and \( \hat{y}_{t+k}^{(i)} \) is the forecast for day \( t+k \) of the \( i \)-th model.

This ensemble aligns with deep learning strategies by employing multiple learners to inform and enhance the decision-making process.  

By leveraging the strengths of multiple specialized transformer models, this novel approach offers several advantages: 

\begin{itemize}
    \item \textbf{Enhanced Accuracy:} Averaging predictions across multiple models reduces the impact of outliers and potential errors from any single model's forecast; 

    \item \textbf{Increased Robustness:} The approach mitigates the risk of overfitting to recent data trends by integrating broader historical perspectives, leading to more stable and reliable predictions;

    \item \textbf{Scalability and Flexibility}: The architecture is designed to be scalable, allowing for the addition of more models to cover longer forecasting horizons or to adapt to changes in the underlying data patterns. 
\end{itemize}
 

\subsection{Dataset Overview and Data Preprocessing}

\textbf{Initial Dataset Overview}

Our study utilizes a comprehensive dataset spanning 28 months, from December 2019 to March 2022, comprising hourly electricity consumption records from 321,561 residential customers. Each record includes a timestamp, customer identification number, and the amount of electricity consumed in the last hour, tallying approximately 20,400 observations per customer.

\textbf{Preliminary Data Cleaning}

To ensure the data's integrity and applicability for residential consumption analysis, we conducted extensive preliminary data cleaning:

\begin{enumerate}
    \item \textbf{Inactive or Discontinued Usage:} We removed records from consumers who did not have readings for a continuous 24-hour period.
    \item \textbf{Unusual Consumption Patterns:} Entries showing unusually low consumption were excluded,  as they were probably from unused households. Similarly, entries exceeding 12 kWh per hour were removed, since these are indicative of industrial or commercial load, rather than residential usage.
\end{enumerate}

This targeted cleaning refined the dataset, allowing us to focus the analysis on genuine residential consumption patterns and reducing the initial dataset to 73,802 active consumer profiles with typical residential energy usage.

\textbf{Advanced Data Preprocessing}

With a refined dataset of consumer profiles, further preprocessing was applied to prepare the data for the detailed analysis carried out by our research:

\begin{enumerate}
    \item \textbf{Noise Reduction and Smoothing:} We employed a simple moving average (SMA) with a 7-day window to smooth the daily electricity consumption data, reducing short-term fluctuations and highlighting long-term trends;
    \item \textbf{Handling Missing Values:} The dataset used in our analysis was well-prepared, with no missing values present. This completeness is a result of the preliminary data cleaning process, which ensured that all data used for the study was continuous and complete, eliminating the need to fill in any gaps;
    \item \textbf{Feature Engineering:} From the basic raw data features, i.e. timestamp, customer identification number and hourly electricity consumption, we derived additional features such as day of the week, month, and year. These details were carefully extracted from the timestamps to show how electricity use changes with the seasons and on a regular basis. This made our analysis much more useful and in-depth by explicitly adding a new context that was only latent in the raw data.
    
    In particular, we extracted time-based features from the 'Date\_Time' column to capture temporal patterns:
    
    \begin{itemize}

    \item Day of the Week ('day\_of\_week'): Indicates the day of the week (0 for Monday, 6 for Sunday). This feature helps the model learn weekly patterns and trends in the data;
    \item Month ('month'): Captures monthly seasonal effects, which can influence the target variable due to factors like holidays or seasonal demand;
    \item Year ('year'): Useful for identifying long-term trends or shifts over different years.
   
   \end{itemize}

From these features, we created the following set of context features: context\_features = ['SMA\_7', 'day\_of\_week', 'month']. We did not include 'year' because it was less relevant, given that our dataset spans only 28 months. 'SMA\_7' represents the consumption readings in kWh. It is included in the context features and also serves as the target variable to be predicted.

Using these three features, we created a contextual feature that represents the correlation between 'SMA\_7', 'day\_of\_week', and 'month'. To achieve this, we applied Principal Component Analysis (PCA) to reduce the dimensionality of the context features from three to one principal component. The benefits of this approach include: (i) simplification: reduces the complexity of the data, making model training more efficient; (ii) variance retention: captures the most significant variance in the data with minimal loss of information; (iii) multicollinearity reduction: mitigates issues arising from highly correlated features. The reduced context data was included as an additional input to the model, potentially enhancing its predictive capabilities by incorporating key contextual information. Our final feature set used for modeling was ['SMA\_7', 'day\_of\_week', 'context\_data\_reduced'].

    \item \textbf{Normalization and Scaling:} The RobustScaler was used on the smoothed consumption data to minimize the influence of outliers. Unlike standard scaling methods, RobustScaler uses the median and interquartile range (IQR), making it less sensitive to outliers. We scaled our 'SMA\_7' feature, which represents the consumption reading values. Since our data are raw and include not only households but also small industries and the overall regional load, the 'SMA\_7' feature may contain outliers due to fluctuations in the time-series data. Using RobustScaler ensures that these outliers do not unduly influence the scaling process.
     
    The StandardScaler was applied to other contextual features to standardize the data. Basically, this consists in computing a standard z-score (subtracting the mean and normalizing by dividing by the standard deviation, scaling to unit variance). We scaled the 'day\_of\_week' feature and the context features. This ensures that the 'day\_of\_week' feature is normalized, making it easier for the model to learn patterns. The 'day\_of\_week' feature is ordinal and ranges from 0 to 6. Standardizing this feature helps in treating all days equally in terms of scale, which will improve the model's performance; 
    \item \textbf{Dimensionality Reduction:} Principal Component Analysis (PCA) was employed not just to reduce the complexity of our feature space, but also to create contextual features that capture essential variations, as previously explained. These PCA-derived features were then added to our model as new variables, enhancing the predictive capability by providing a condensed representation of critical data dimensions, which improves the effectiveness of our analysis.

\end{enumerate}

\textbf{Clustering for Detailed Analysis}
The cleaned and transformed data were used to perform clustering on the selected 73,802 consumer profiles. This step involved grouping consumers into distinct categories based on their energy consumption behaviours, which is fundamental for subsequent tailored analysis and for developing specific energy conservation strategies.

\textbf{Data Modeling}

The thoroughly preprocessed and clustered data provided a robust foundation for deploying sophisticated predictive models. These steps ensured high-quality data inputs, enabling more reliable and accurate forecasts of residential energy consumption patterns.

\subsection{Training the model}

In this section, we provide details related to the training of our Just In Time transformer model, whose code is freely available\footnote{https://github.com/cafaro/JIT-Transformer}. The number of parameters to be learned is slightly more than 125 million. These have been learned by using the dataset, which includes hourly measurements related to 73,802 customers. We have aggregated the available measurements by computing the average daily consumption so that the dataset provides for each user his/her mean consumption for a period spanning 852 days.

The dataset itself has been split as follows in order to train the model: 80\% for training, 10\% for validation, and 10\% for testing. The chosen loss function and the optimizer algorithm were, respectively, Mean Squared Error (MSE) and Adam. For the systematic and efficient selection of the optimal hyperparameters, we utilized Optuna, a hyperparameter optimization framework. Optuna employs Bayesian optimization techniques to explore the parameter space and has been integral in determining the best set of parameters for our model on the basis of the performance metrics defined in our experimental setup.

\begin{figure*}[h!]
    \centering
    \includegraphics[width=\textwidth, height=0.35\textheight]{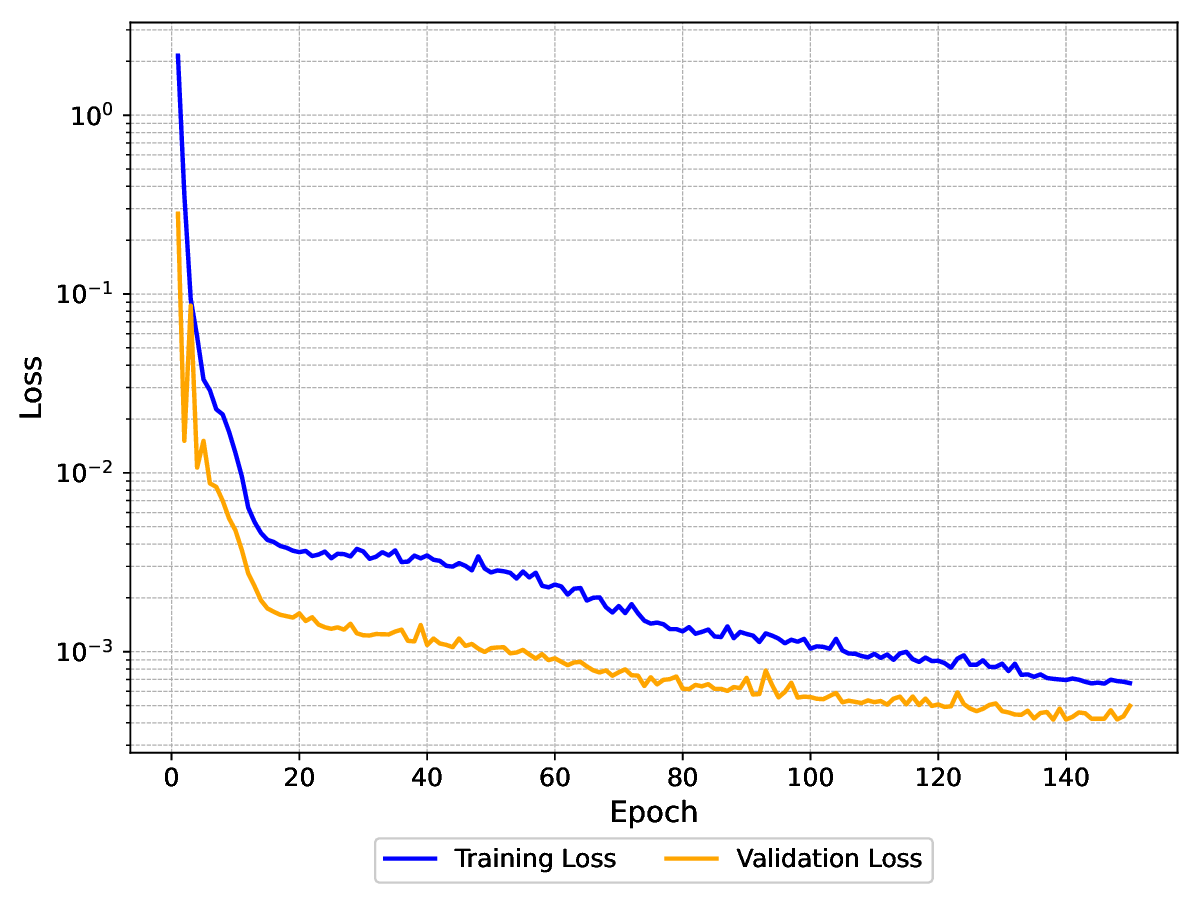} 
    \caption{JITtrans training and validation losses.}
    \label{losses}
\end{figure*}

The model has been trained on the italian Leonardo supercomputer\footnote{https://leonardo-supercomputer.cineca.eu} for 150 epochs with a batch size equal to 64 and a learning rate initially set to 0.0001, using one NVIDIA Ampere A100 Tensor Core GPU equipped with 64 GB HBM2e of memory and NVLink 3.0 (200 GB/s). The learning rate and other hyperparameters were optimized through Optuna to maximize the model's performance on the validation set. The training process required about 274 seconds. Moreover, we have also verified that the model, besides on the data center NVIDIA Ampere A100 Tensor Core GPU equipped with 64 GB of RAM, can be trained on consumer grade GPUs, by training it on a NVIDIA RTX 4090 GPU with 24 GB of RAM in about 201 seconds. The NVIDIA RTX 4090 is a more recent GPU than the Ampere A100, it is cheaper and slightly faster. Figure \ref{losses} shows the training and the validation loss.

\section{Experimental Results}
\label{results}

Here, we present and discuss the results of our study, including the determination of the optimal number of clusters, the visualization of the clustering results, the performance comparison across different models, and a detailed analysis of the temporal predictions.

\subsection{Cluster Determination and Visualization}
\begin{figure*}[h]
 
    \begin{subfigure}[b]{0.5\textwidth}
        \adjustbox{valign=t}{\includegraphics[width=\textwidth]{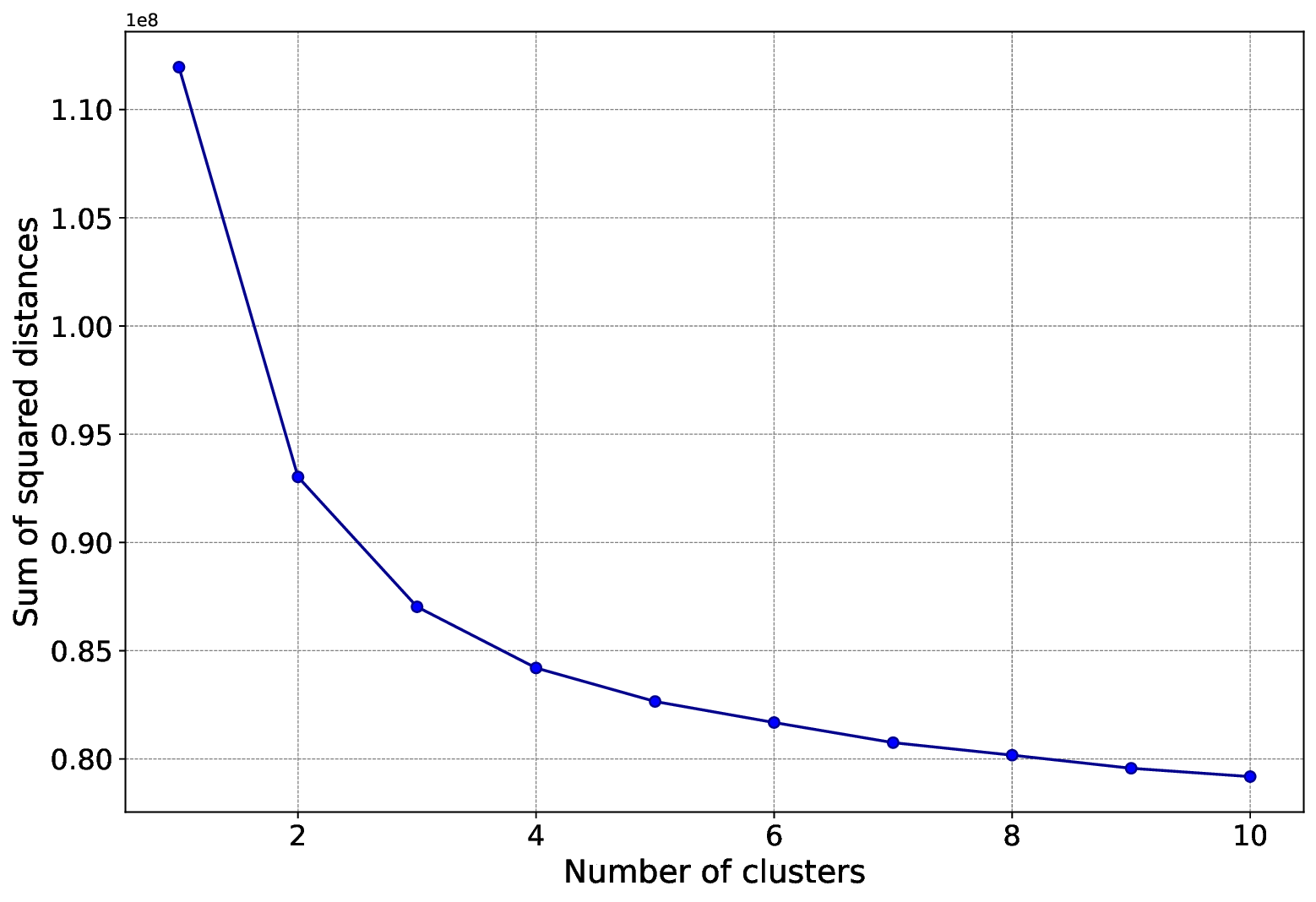}}
        \caption{Elbow method to establish the optimal number of clusters.}
        \label{fig:elbow}
    \end{subfigure}
    \qquad
    \begin{subfigure}[b]{0.5\textwidth} 
        \adjustbox{valign=t}{\includegraphics[width=\textwidth]{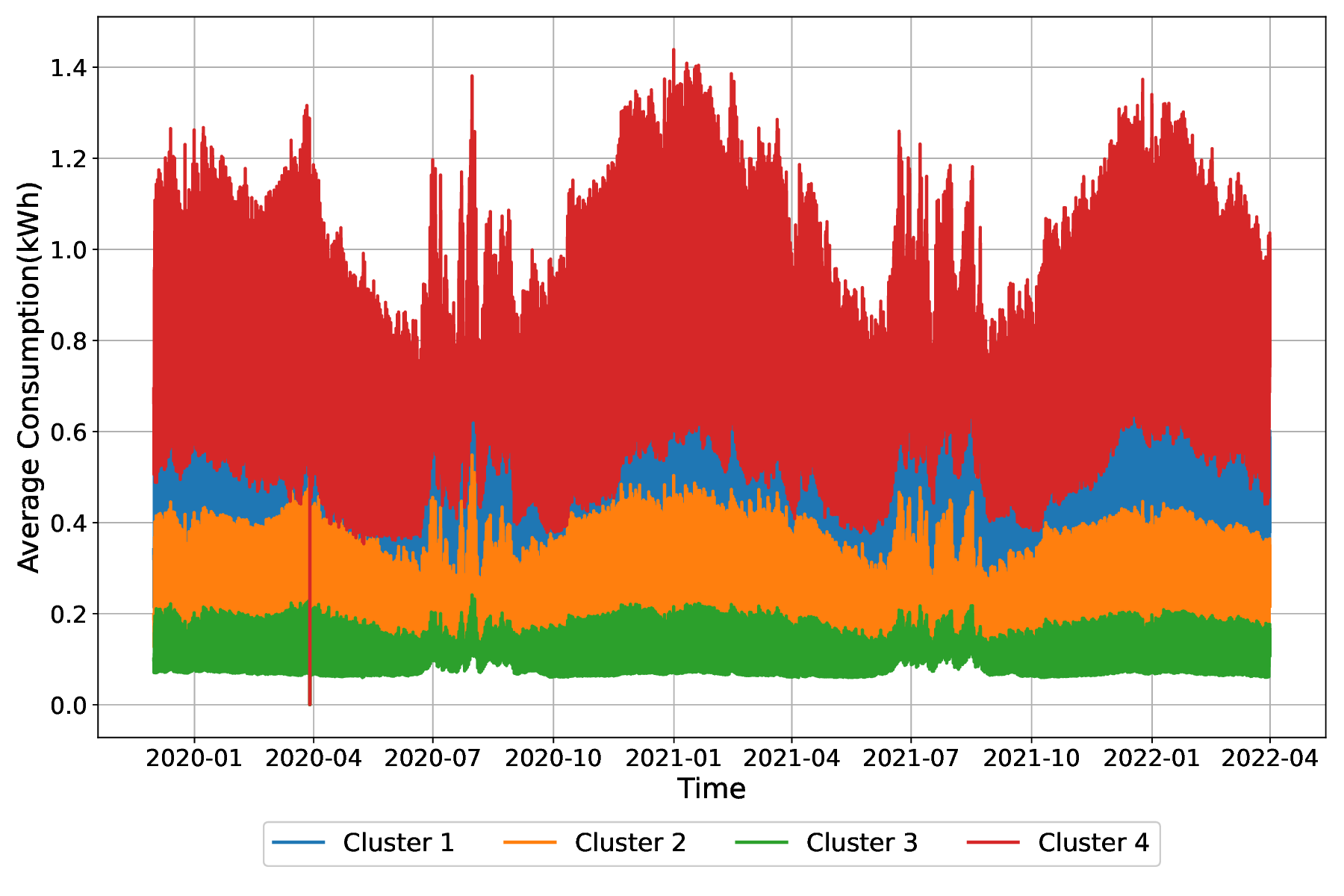}}
        \caption{Average consumption patterns for each cluster.}
        \label{fig:avg_consumption}
    \end{subfigure}
    \caption{Clustering and average consumption patterns.}
    \label{fig:elbow_avg_consumption}
\end{figure*}

The Elbow method was employed to determine the optimal number of clusters for $k$-means. Figure \ref{fig:elbow} shows the elbow plot, which indicates that four clusters are optimal, as the point where adding more clusters does not significantly improve the variance explained by the model. The clustered data visualization shows distinct groupings of customers with similar consumption patterns. This clustering simplifies the modelling process and enhances the efficiency of the forecasts. Figure \ref{fig:avg_consumption} clearly depicts the average consumption patterns for each cluster, demonstrating the variance in energy usage behaviors. For instance, we can observe that the average consumption in Cluster 4 significantly differs from that in Cluster 3, indicating distinct energy usage behaviors between the two groups. This differentiation is also apparent across all clusters, each exhibiting its unique average consumption profile. In turn, the high variability discovered among the clusters directly influences the model design. The clustering allowed us to tailor the forecasting model to handle different consumption behaviors more effectively. This design choice ensures the model’s robustness and adaptability to the diverse energy consumption patterns.

\subsection{Evaluation Methodology}

The predictive accuracy of the models was quantified using the average error metric. This metric, specifically the Mean Absolute Error (MAE), is computed as follows:

\begin{equation}
	\text{MAE} = \frac{1}{n} \sum_{i=1}^{n} |P_i - A_i|,
\end{equation}

where:
\begin{itemize}
    \item $n$ is the number of data points;
    \item $P_i$ represents the predicted values of electricity consumption;
    \item $A_i$ represents the actual values of electricity consumption.
\end{itemize}

The absolute differences between predicted and actual values are summed up and then divided by the total number of observations, providing a straightforward measure of model accuracy in kWh.

\subsection{Performance Comparison Across Clusters}

\begin{figure*}[]
\centering
\includegraphics[width=\textwidth]{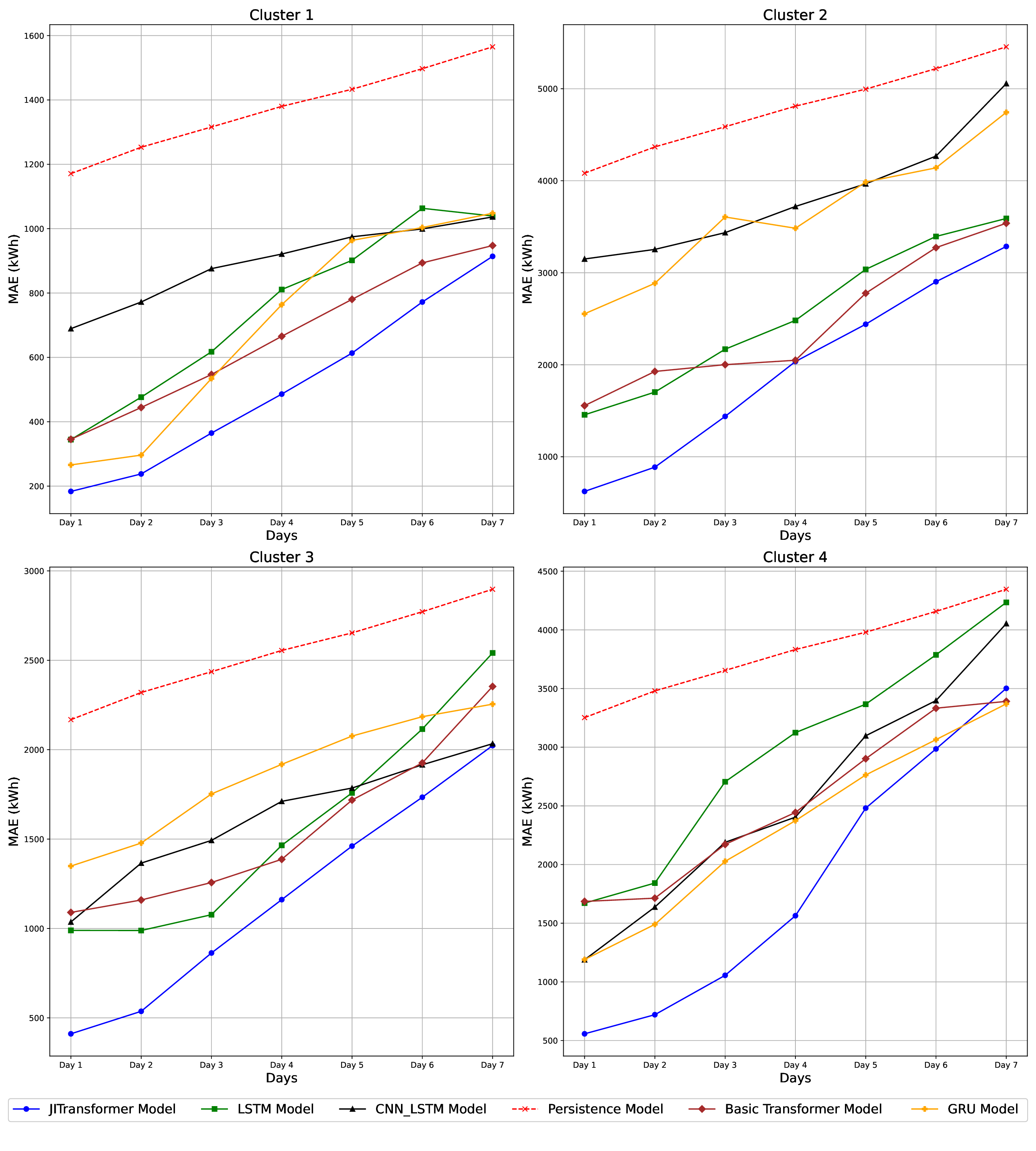}
       \caption{Comparison of average error per day across clusters.}
       \label{cluster-comparison}
\end{figure*}

The effectiveness of this model is compared against a basic transformer, a GRU model, a CNN-LSTM model, a LSTM model, and the persistence model\footnote{The persistence model is a standard benchmark model, that works assuming that the future value of a time series is computed under the assumption that nothing changes between the current time and the forecast time.} across the four clusters. Figure \ref{cluster-comparison} provides a comparison of average errors per day for each cluster. The performance analysis clearly show the JITtrans model's superior ability to handle diverse energy usage patterns. 

\begin{itemize}
    \item Cluster 1: this cluster exhibits the least variability in errors, suggesting that households within this cluster have more predictable energy consumption patterns, due to similar lifestyle or appliance usage;
    \item Clusters 2 and 3: these clusters showed higher variability and errors, reflecting the challenges posed by more dynamic and diverse energy usage behaviours. The JITtrans model was particularly effective in these clusters, reducing prediction errors and demonstrating its capacity to adapt to complex scenarios;
    \item Cluster 4: this cluster, which is the most heterogeneous in terms of energy usage, exhibits the highest errors across all models, emphasizing the challenges in forecasting for highly variable consumer behaviours. Despite this difficulty, our JITtrans model outperformed other techniques, showcasing his robustness.
\end{itemize}

\subsection{Temporal Analysis of Predictions}

\begin{figure*}[]
    \centering
    \includegraphics[width=\textwidth]{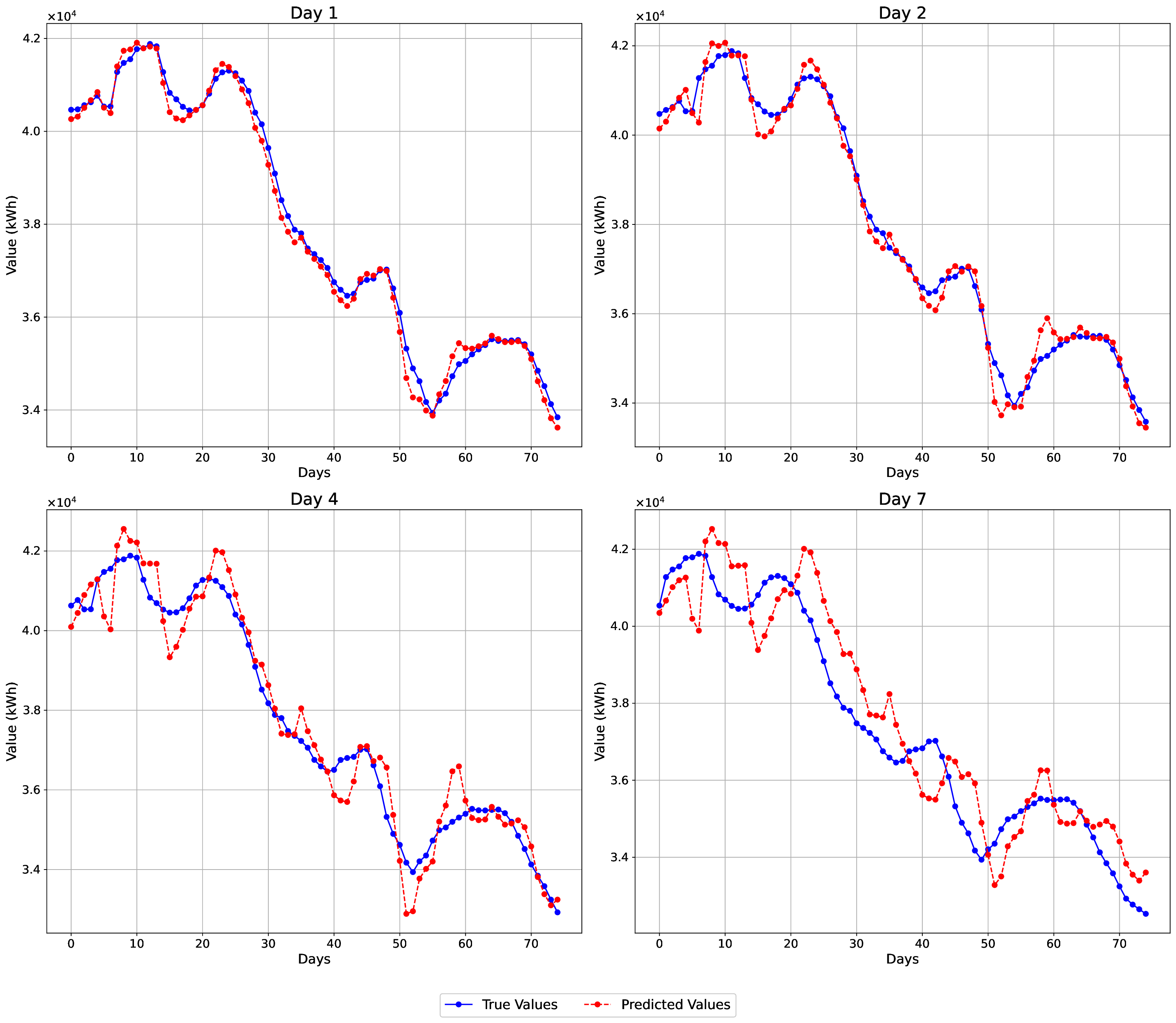}
       \caption{Comparison between the prediction and the ground truth varying the lead time.}
       \label{fig:temporal_predictions_days}
\end{figure*}

\begin{figure*}[]
    \centering
   \includegraphics[width=\textwidth]{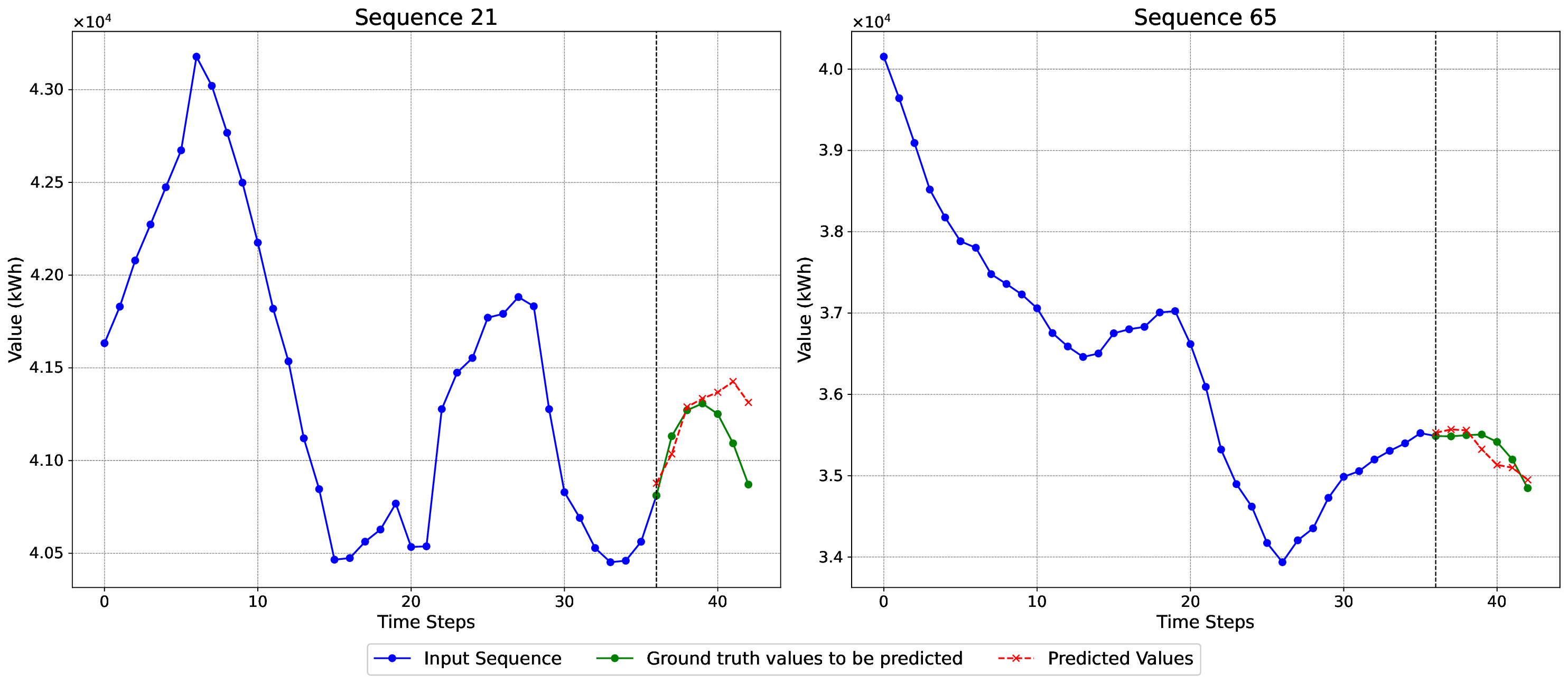}
    \caption{Comparison between the prediction and the ground truth.}
    \label{fig:temporal_predictions_sequences}
\end{figure*}

Figures \ref{fig:temporal_predictions_days} and \ref{fig:temporal_predictions_sequences} depict the true vs. predicted values for some sequences and days across different clusters, highlighting the JITtrans model adeptness at closely mirroring actual consumption patterns and minimizing discrepancies. Particularly on days 1, 2 and 3, the model exhibits high accuracy, which is crucial for effective load management. A slight underestimation can be noted on day 7 and only for cluster 4. In this case the GRU and the basic transformer model perform just slightly better. However, it is worth noting here that (i) the observed behaviour is expected, i.e. it is normal for the prediction accuracy to decrease as the forecasting horizon increases, and (ii) the user (in our case the stakeholder is the power company) is not really interested in accurately predicting the load on such a distant forecasting horizon: the main focus is on predicting the load for the next day. 

Regarding specific sequence predictions, the examination of some randomly selected sequences - Sequence 21 and Sequence 65 depicted in Figure \ref{fig:temporal_predictions_sequences} - showed high initial accuracy with slight deviations towards the end. Again, this is the expected behaviour. We observed stable pattern prediction with minor discrepancies, reinforcing the model's robustness and reliability.

This portion of our analysis focuses on selected days as representative examples (owing to the impossibility of presenting detailed daily results for all four clusters and the fact that those results are essentially similar). The observed patterns, consistent across different clusters, reinforce the robustness of our modelling approach and confirm the JITtrans model's capability to adapt to and accurately forecast diverse energy usage behaviours.

\subsection{Discussion}
The JITtrans model demonstrated a significant improvement over traditional forecasting models such as LSTM, CNN\_LSTM, and GRU in predicting household electricity consumption. This enhanced performance is primarily due to the transformer's sophisticated capacity to simultaneously process and analyze complex temporal dependencies and patterns within clustered data. Integrating customers with similar consumption behaviours into clusters not only streamlined the forecasting process but also significantly enhanced overall efficiency. This approach eliminated the need for creating separate models for each household, thus addressing the impracticality of such an endeavour in real-world applications. The JITtrans model greatly improves the accuracy in forecasting, allowing utility companies to optimize energy distribution strategies effectively. By precisely predicting consumption patterns, these utilities can manage demand fluctuations more efficiently, seamlessly integrating renewable energy sources, and optimizing their energy supply chains. Such advancements in predictive capabilities are crucial for adapting to the changing dynamics of global energy needs, promoting sustainable practices, and reducing wastage in energy distribution, marking a significant step forward in the development of smart energy management solutions.

\section{Conclusion}
\label{conclusions}

In this paper we introduced a novel approach to energy load forecasting, based on our JITtrans model specifically designed for accurate energy consumption forecasting. We enhance traditional transformers by implementing specific training for each day in the prediction horizon, where a dedicated model averages the prediction errors of previous models to predict each day. This design allows for more effective and accurate time series forecasting by reducing cumulative errors and leveraging information from previous predictions.

Our model demonstrated superior performance in various numerical benchmarks, outperforming state-of-the-art models in accuracy and robustness. We attribute this improvement to the model's ability to explore and utilize information within each token from previous predictions, effectively mitigating prediction errors over time.

From a theoretical perspective, future research will delve into refining this model by investigating novel error reduction strategies and optimization techniques. From a methodological standpoint, we aim to extend the application of our model to additional time series forecasting domains, testing its adaptability and effectiveness. From a practical viewpoint, we envisage the integration of external variables such as climatic conditions and economic data to further refine the forecasting capabilities of our model. Planned future developments include:

    \begin{itemize}
    \item  \textit{Real-Time Prediction:} Implement the enhanced model in a real-time energy management system to continuously forecast energy consumption;

    \item \textit{Feedback Loop:} Setup a feedback loop to update the model with new data, including changes in user behaviour, to keep the predictions accurate over time;
    \item \textit{Regional datasets:} Assess the behaviour of our model on several datasets characterized by different geolocations.
	\end{itemize}

\section*{Acknowledgment}
We are grateful to Prof. Massimo Tavoni and Prof. Jacopo Bonan (Politecnico di Milano, Milan, Italy) for providing the dataset used to assess our Just In Time Transformer model.

\clearpage

\bibliographystyle{plain}
\bibliography{references}

\end{document}